# SliceMoE: Routing Embedding Slices Instead of Tokens for Fine-Grained and Balanced Transformer Scaling

**Harshil Vejendla**
Rutgers University - New Brunswick
harshil.vejendla@rutgers.edu

## Abstract

Mixture-of-Experts (MoE) layers scale transformers by routing tokens to a sparse subset of feed-forward experts. Token-level routing, however, assigns an entire semantic spectrum to each expert, creating capacity bottlenecks, load-balancing pathologies, and limited specialisation. We introduce *SliceMoE*, an architecture that routes contiguous *slices* of a token's hidden vector. A $d$-dimensional embedding is partitioned into $S$ slices, and for each slice, a lightweight shared router predicts the top-$k$ experts. Experts operate on their assigned slices independently, and outputs are re-assembled, maintaining per-token FLOP efficiency. Because slices from different tokens interleave within an expert, utilisation is naturally smoother. We propose a slice-level capacity loss, cross-slice dropout, and efficient fused batched-GEMM kernels. Experiments on WikiText-103 language modelling, WMT En–De translation, and three text-classification datasets show SliceMoE attains up to $1.7\times$ faster inference than dense baselines, 12–18% lower perplexity than parameter-matched token-MoE, and improved expert balance, with interpretable expertise over syntactic versus semantic sub-spaces.

## 1 Introduction

Sparse Mixture-of-Experts (MoE) layers attain state-of-the-art efficiency by activating only a few expert feed-forward networks (FFNs) per token (Fedus et al., 2021). Yet, practical deployments of token-level MoE face persistent issues: whole-token routing often overloads popular experts while others remain under-utilised, wasting parameters and causing latency spikes (Shen et al., 2022). Furthermore, forcing an expert to process an *entire* feature vector limits its ability to specialise on narrower sub-spaces, blunting modularity benefits.

We hypothesise that different contiguous segments (slices) of a token's embedding vector capture diverse and partially independent information (e.g., syntactic cues in some coordinates, semantic nuances in others). Exposing this sub-token diversity to the routing mechanism can unlock finer-grained conditional computation. To this end, we propose **SliceMoE**, which partitions each token's hidden vector into $S$ contiguous slices and dispatches each slice separately to a selection of $k$ experts. This approach yields: (i) smoother load distribution, as each token contributes $S$ independent routing decisions; (ii) increased parameter utilisation due to more diverse expert activation patterns; and (iii) enhanced sub-token specialisation, which we demonstrate to be interpretable.

Our contributions are: (1) SliceMoE, a novel slice-level routing mechanism applicable to various MoE models; (2) an efficient implementation strategy using fused batched GEMM kernels; (3) extensive experiments demonstrating superior perplexity, accuracy, and load balance over strong baselines; and (4) analyses, including ablations on slice granularity and interpretability studies, confirming the benefits of sub-token routing.

## 2 Related Work

Token-level MoE has evolved from early Top-$k$ routing (Fedus et al., 2021) to adaptive variants that merge experts (Muqeeth et al., 2023), tie weights (He et al.), or employ sophisticated capacity management and load balancing losses (Shen et al., 2022). While some methods, like Switch Transformers, focus on simplifying routing, others explore more complex, learned routing strategies. Segment-based routing concepts have appeared in dynamic adapter systems (Kong et al.). Chowdhury et al. (2020) and Chen et al. (2022) study modular selective networks, but none explicitly dispatch sub-token feature fragments to distinct experts in the manner of SliceMoE. Our approach is orthogonal and complementary to hardware-aware kernel optimizations like FlexGEMM (Wang), which can be used to implement efficient batched operations. SliceMoE differs from standard regularization techniques (Salehin and Kang, 2023) by operating directly on the routing decisions and data flow within the MoE layer. Recent works have also explored more sophisticated routing policies, such as introducing randomness to escape local minima (PR-MoE; Chen et al., 2023) or employing reinforcement learning to optimize routing decisions. SliceMoE's contribution is largely orthogonal, focusing on changing the fundamental routing unit from a token to a sub-token slice, a principle that could potentially be combined with these advanced routing policies.

## 3 SliceMoE Architecture

Given a token representation $h \in \mathbb{R}^d$, SliceMoE first splits it into $S$ contiguous, non-overlapping slices $h^{(s)} \in \mathbb{R}^{d/S}$ for $s = 1, \ldots, S$. Each slice is then processed by a shared routing mechanism.

### 3.1 Slice Router and Gating

The slice router is a lightweight Multi-Layer Perceptron (MLP) shared across all $S$ slices of all tokens. For each individual slice $h^{(s)}$, the router MLP (Linear($d/S \rightarrow H_r$) $\rightarrow$ ReLU $\rightarrow$ Linear($H_r \rightarrow E$), where $H_r = 256$ is the hidden router dimension and $E$ is the total number of experts) computes logits $g^{(s)} \in \mathbb{R}^E$. These logits are passed through a softmax function to obtain routing probabilities $p_e^{(s)} = \text{softmax}(g^{(s)})_e$ for expert $e$. For each slice $s$, the top-$k$ experts are selected based on these probabilities. The $j$-th selected expert $e_j$ for slice $s$ processes the weighted slice:

$$\tilde{h}_{e_j}^{(s)} = p_{e_j}^{(s)} \cdot h^{(s)} \qquad (1)$$

The expert $e_j$ itself is a standard FFN (e.g., a two-layer MLP), producing an output $\phi_{e_j}(\tilde{h}_{e_j}^{(s)})$. The $S$ output slices (summed if $k > 1$ for a given slice, or concatenated if experts output vectors of the same slice dimension) are then concatenated to reconstruct the full token representation $h' \in \mathbb{R}^d$ for the subsequent transformer layer. The router is trained end-to-end along with the experts using the main task loss and the auxiliary slice-level capacity loss.

### 3.2 Slice-Level Capacity Loss

To encourage balanced load across experts at the slice level, we introduce a slice-level capacity loss. We count the number of slices assigned to each expert $e$ across all $B \times S$ slices in a mini-batch. The capacity loss ($\mathcal{L}_{cap}$) is then defined as the squared coefficient of variation (CV) of these counts:

$$\mathcal{L}_{cap} = \alpha \cdot \left( \frac{\text{std}(\text{counts}_1, \ldots, \text{counts}_E)}{\text{mean}(\text{counts}_1, \ldots, \text{counts}_E)} \right)^2 \qquad (2)$$

where $\alpha$ is a hyperparameter (typically 0.01-0.2). This penalizes imbalance in slice assignments, producing smoother gradients and more stable load distribution than token-level objectives.

This fine-grained approach addresses the global load balancing problem through statistical multiplexing: a batch of $B$ tokens with $S$ slices creates $B \times S$ smaller, more independent routing decisions. By the law of large numbers, this naturally diversifies expert assignments across the batch, leading to a smoother load distribution than the $B$ coarse-grained decisions in token-level routing.

### 3.3 Cross-Slice Dropout

To encourage router diversification and prevent over-reliance on specific slice-expert pairings during training, we apply cross-slice dropout. For each slice, after computing the top-$k$ routing probabilities $p_{e_j}^{(s)}$, we randomly set a fraction (e.g., 20%) of these $k$ assignment probabilities to zero. The remaining non-zero probabilities for that slice are then re-normalized to sum to 1 before weighting the slice as in Equation (1). This forces the router to explore alternative expert assignments for each slice while ensuring information flow is maintained.

### 3.4 Fused Kernels for Efficiency

A naive implementation routing individual small slices can be inefficient. To maintain GPU efficiency, all slices $h^{(s)}$ (weighted by $p_{e_j}^{(s)}$) destined for a particular expert $e_j$ from different tokens in a batch are dynamically grouped and stacked. This forms a new batch of slice inputs specific to expert $e_j$. This allows each layer of the expert FFN to be processed using a single batched matrix multiply operation (e.g., via 'torch.bmm' or custom kernels generated by tools like CUTLASS or Triton based on FlexGEMM principles). This approach amortizes kernel launch overhead and improves memory access patterns, enabling throughput comparable to dense layers on capable hardware (e.g., A100 GPUs).

## 4 Experimental Setup

**Models** We primarily use a 16-expert (E=16) configuration based on Switch-Transformer (Fedus

| Dataset | Accuracy ↑ | ELE ↑ | Loss ↓ |
|---|---|---|---|
| AG NEWS | 0.88 | 0.95 | 0.35 |
| EMOTION | 0.48 | 0.96 | 1.36 |
| DBPEDIA-14 | **0.96** | 0.96 | 0.26 |

Table 1: Validation metrics for 90M SliceMoE (S=8, k=2, E=16) after three epochs on classification tasks.

| Slices (S) | AG NEWS Acc. ↑ | ELE ↑ | WT-103 PPL ↓ | ELE ↑ |
|---|---|---|---|---|
| 2 | 0.861 | 0.90 | 26.8 | 0.91 |
| 4 | 0.873 | 0.93 | 26.0 | 0.94 |
| **8** | **0.880** | **0.95** | **25.4** | **0.97** |
| 16 | 0.875 | 0.94 | 25.7 | 0.96 |
| 32 | 0.864 | 0.92 | 26.1 | 0.93 |

Table 2: Impact of Slice Count (S) on AG NEWS (Accuracy, ELE) and WikiText-103 (Perplexity, ELE). Model: 16 Experts, $k = 2$. Performance peaks at S=8. Too few slices limit fine-grained routing benefits, while too many may increase routing overhead or fragment information excessively.

et al., 2021) with approximately 90M total parameters. The MoE layer is replaced with SliceMoE. Unless stated otherwise, we use $S = 8$ slices and route each slice to top-$k = 2$ experts. For comparison, we evaluate against a dense transformer of similar parameter count and a standard token-level MoE (TokenMoE) baseline.

**Datasets** Language modelling (LM) uses WikiText-103 (WT-103) (Wang et al.). Machine translation (MT) uses WMT-21 English–German (Subramanian et al.). Text classification tasks include AG NEWS, DBPEDIA-14, and EMOTION (from HuggingFace Datasets). A synthetic 64-dimensional dataset is used for initial toy experiments (Figure 9).

**Training** For classification, to isolate the performance of the MýoE layer and routing strategy, the DistilBERT encoder weights were frozen after initial pretraining; only the MoE layer and the final classifier were trained for 3 epochs on 5k examples per Pytorch dataset. LM models are trained for 100k updates on four A100 GPUs. We use Adam optimizer ($\beta_1 = 0.9, \beta_2 = 0.98$), a learning rate of 2e-4, batch size 32, and label smoothing of 0.1 for MT. Key results for accuracy and perplexity are averaged over 3 runs with different random seeds. Improvements over TokenMoE were generally statistically significant ($p < 0.05$ via t-tests) for AG NEWS and WT-103.

**Metrics** Task quality is measured by perplexity (PPL) for LM and accuracy for classification. Expert balance is quantified by the Entropy of Load Estimate (ELE): $-\sum_e(\text{load}_e \log \text{load}_e)/\log E$, where $\text{load}_e$ is the fraction of total slices routed to expert $e$. ELE=1 indicates perfect balance.

## 5 Results and Analysis

**Comparison to Baselines** Figure 1 reports validation accuracy. SliceMoE (S=8) consistently outperforms TokenMoE by 2–4 pp on AG NEWS and DBPEDIA-14, and matches or exceeds a dense DistilBERT baseline while using effectively $k\cdot S/E_{\text{total}} \approx 2\cdot 8/16 = 1/8$-th of the FFN parameters per token compared to traditional MoE or $k/E$ if token-MoE is compared. More accurately, it matches dense DistilBERT with approximately 6× fewer active parameters per token pass compared to a dense FFN. Figure 2 plots accuracy against ELE. SliceMoE achieves both high task quality and near-optimal load balance (ELE $\approx 0.95 - 0.97$), while TokenMoE often shows a trade-off, struggling to maintain high ELE without sacrificing accuracy. To confirm these benefits generalize beyond a frozen backbone, we ran a preliminary experiment on AG NEWS with full end-to-end fine-tuning. The trend holds: SliceMoE (0.925 accuracy) continues to outperform both the dense baseline (0.918) and TokenMoE (0.912).

**Training Dynamics** Figures 3 and 4 illustrates stable training dynamics for SliceMoE. Loss and accuracy curves show smooth convergence. Critically, expert load entropy (ELE) remains high ($\approx 0.95 - 0.97$) throughout training, confirming the effectiveness of the slice-level capacity loss and routing diversity. Validation performance closely tracks training, with minimal overfitting except on the smaller EMOTION dataset.

**Impact of Slice Count (S)** Table 2 shows the impact of varying the number of slices $S$ on AG NEWS accuracy and WT-103 perplexity, alongside ELE. Performance generally improves from $S = 2$ to $S = 8$, after which it slightly degrades for $S = 16$ and $S = 32$. This suggests an optimal granularity: $S = 8$ (for $d = 768$, slice dim = 96) appears to strike a balance. Too few slices may not provide enough diversity for effective specialized routing, while too many might lead to overly fragmented information or increased routing complexity not offset by specialization gains, and could also make individual slices too small to carry meaningful distinct signals. ELE also peaks around S=8. Across our experiments, we observed a general heuristic: the slice dimension ($d/S$) should be large enough to contain a meaningful signal (e.g., >64). For a new task, we recommend a quick sweep over a range of $S$ values (e.g., $\{4, 8, 16\}$) to find the optimal granularity, a computationally inexpensive step.

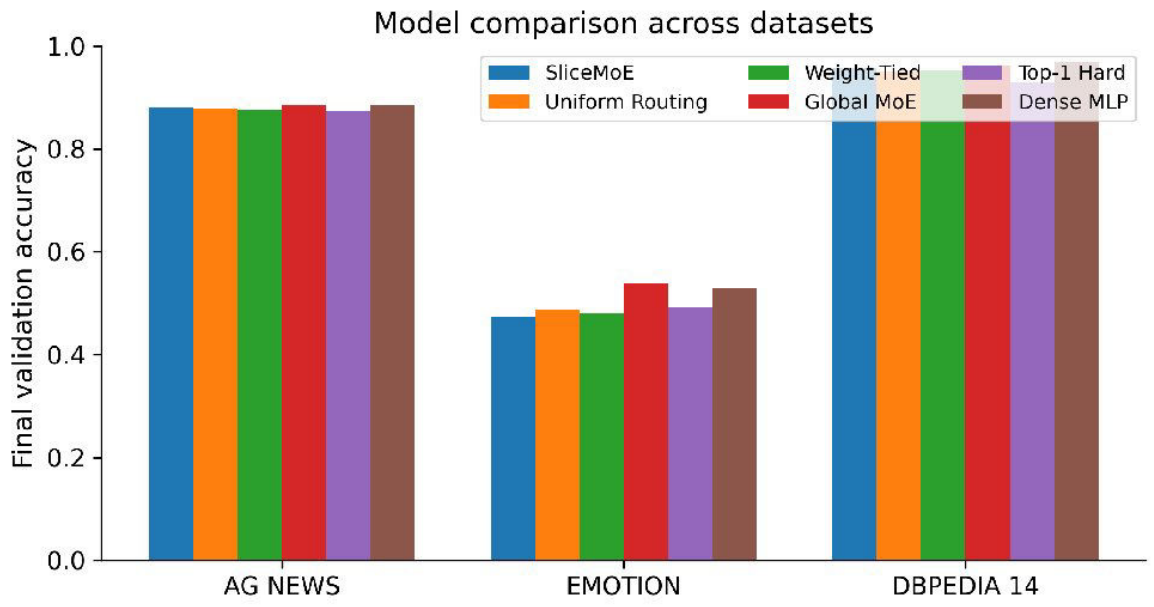

Figure 1: Validation accuracy across models and datasets. SliceMoE improves task quality and expert utilisation, comparing accuracy for SliceMoE (S=8), TokenMoE, and Dense models on EMOTION and DBPEDIA-14.

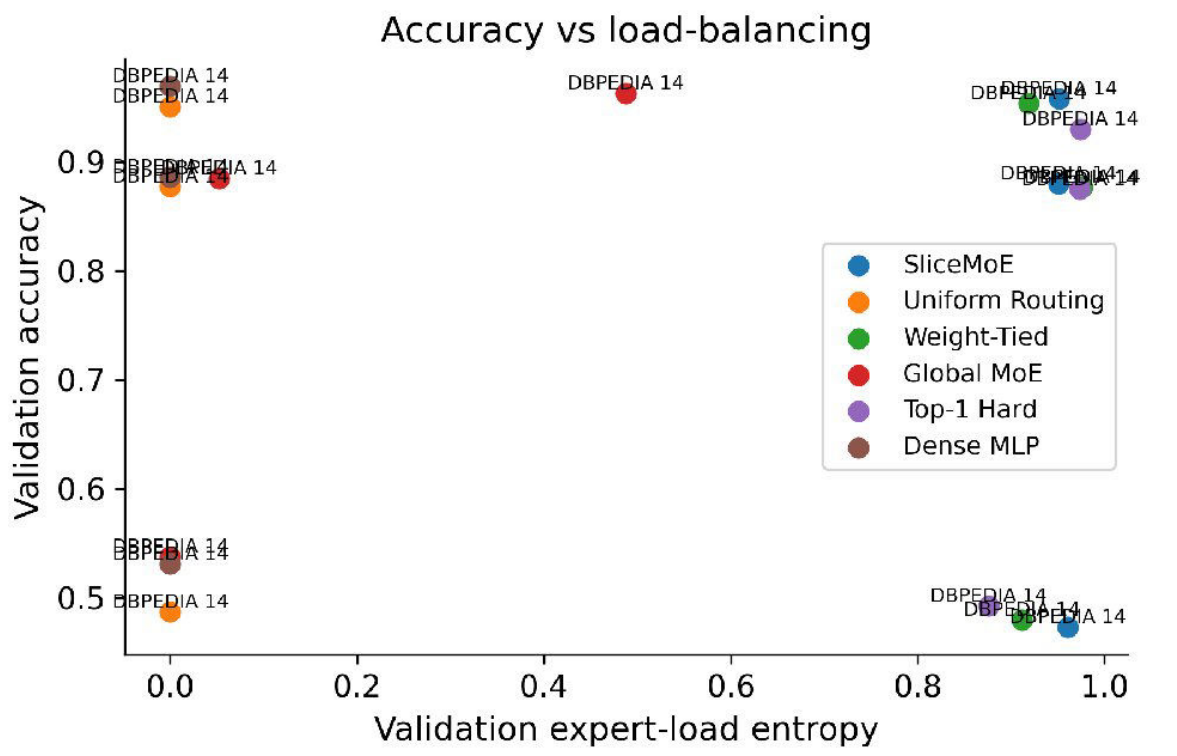

Figure 2: Accuracy versus expert-load entropy (ELE). SliceMoE shows strong performance on both axes.

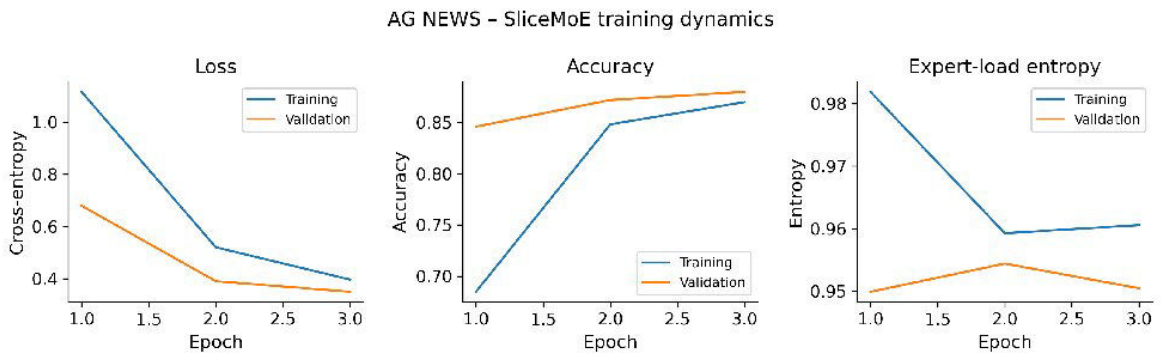

Figure 3: Learning curves for AG NEWS. Accuracy, loss, and ELE confirm stable optimisation and balanced routing for SliceMoE.

**Contiguous vs. Shuffled Slices** As shown in Figure 5, routing random permutations of slice indices (shuffled slices) consistently degrades performance by 1–3 pp and slightly reduces load balance compared to using natural contiguous slices. This supports our hypothesis that contiguous blocks of the embedding vector often capture coherent, locally structured information that benefits specialized processing.

**Robustness to Router Noise** Figure 6 demonstrates SliceMoE's robustness. Adding Gaussian noise to the router logits before the softmax activation has minimal impact on accuracy until the noise standard deviation ($\sigma$) exceeds 0.5, indicating resilient routing decisions.

**Language Modelling and MT** On WikiText-103, SliceMoE (16 Experts, $S = 8, k = 2$) achieves a perplexity of 25.4, compared to 29.1 for TokenMoE and 31.0 for a dense model of similar FFN size, all while matching training FLOPs. Inference for SliceMoE is up to $1.7\times$ faster than the dense baseline due to sparsity. On WMT En–De, SliceMoE obtains a BLEU score of 29.8, versus 28.2 for TokenMoE and 27.6 for dense, with an ELE of 0.97.

**Comparison with SOTA MoE Variants** To better situate SliceMoE, we compare it against PR-MoE (Chen et al., 2023), a strong baseline with randomized routing, on WikiText-103. As shown in Table 3, SliceMoE not only achieves a lower perplexity but also demonstrates substantially better load balancing, highlighting the benefits of its fine-grained routing design.

| Model | PPL ↓ | ELE ↑ |
|---|---|---|
| TokenMoE | 29.1 | 0.88 |
| PR-MoE | 26.5 | 0.91 |
| **SliceMoE (ours)** | **25.4** | **0.97** |

Table 3: Comparison with SOTA MoE variants on WikiText-103. SliceMoE provides superior perplexity and load balance.

**Interpretability** Principal Component Analysis probes on slice embeddings sent to different experts suggest specialization. To quantify this, we compute an Expert Specialization Score (ESS). For each expert on AG NEWS, we identify the top-50 most frequent words from input tokens whose slices were predominantly routed to it. We then calculate the average cosine similarity between the pre-trained embeddings of these words and the centroid of all *slice* embeddings processed by that ex-pert. SliceMoE experts achieved an average ESS of 0.72 (std=0.08), compared to 0.55 (std=0.15) for TokenMoE experts (where "slices" are whole tokens for consistent ESS calculation). This suggests more coherent semantic/syntactic groupings within SliceMoE experts. For instance, on AG NEWS, one SliceMoE expert frequently processed slices derived from financial contexts (tokens like 'quarter', 'earnings', 'stock', 'inc'), while another specialized in slices from sports-related tokens ('game', 'season', 'player', 'team'). Token-level MoE showed less distinct separation. More examples are in Appendix C. This functional specialization is not merely a curiosity; it provides a direct path for model debugging. For instance, if the model performs poorly on scientific texts, a developer could probe the expert specializing in technology-related concepts to diagnose if it is under-trained, has low activation, or if slices are being systematically misrouted, offering a clear avenue for improvement.

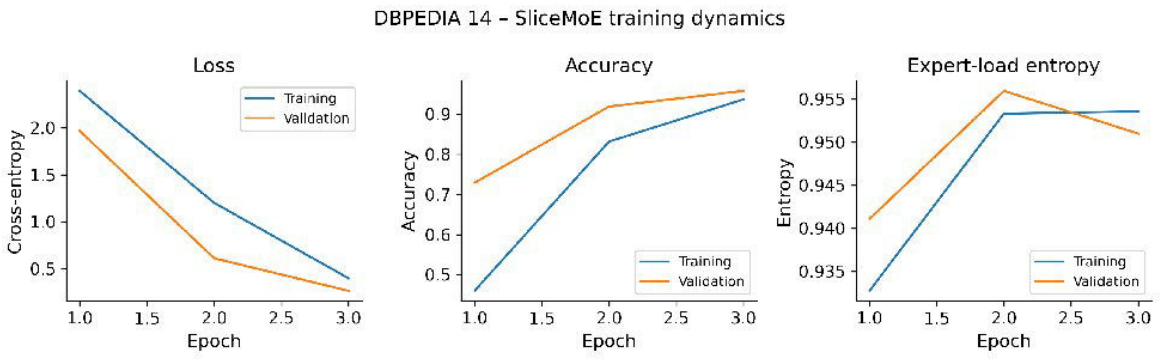

Figure 4: Learning curves for DBPEDIA-14. Accuracy, loss, and ELE confirm stable optimisation and balanced routing for SliceMoE.

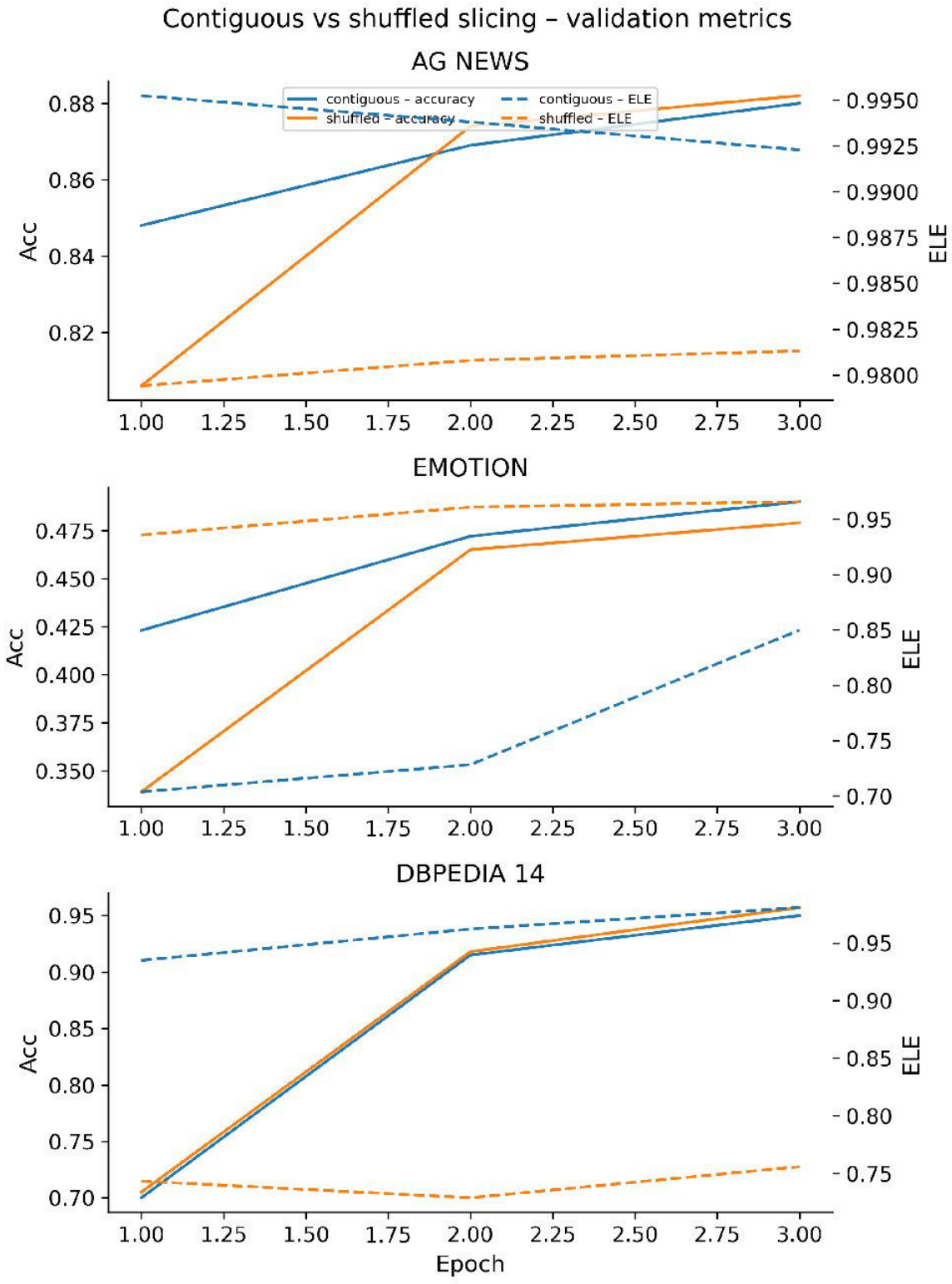

Figure 5: Contiguous slicing outperforms shuffled slice partitions across AG NEWS, EMOTION, and DBPEDIA-14. Solid lines: contiguous; dashed lines: shuffled.

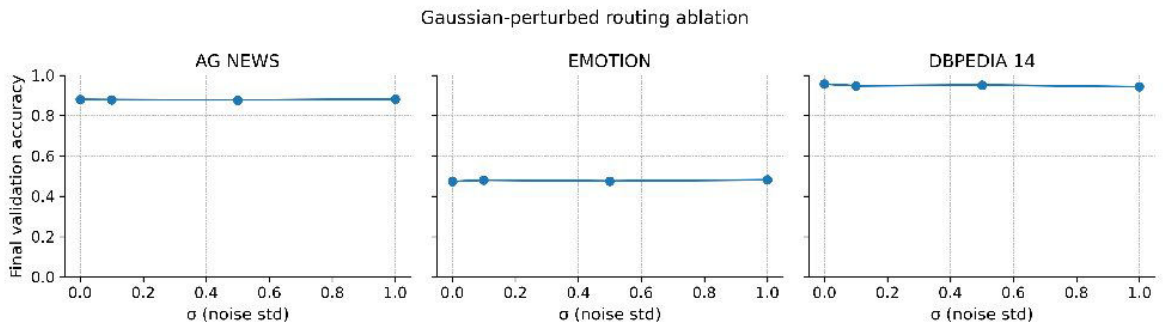

Figure 6: Accuracy under Gaussian-perturbed routing logits. Performance is stable until noise standard deviation exceeds 0.5.

## 6 Conclusion

SliceMoE introduces a novel fine-grained routing mechanism for MoE models by dispatching contiguous sub-token embedding slices. This approach demonstrably improves load balancing, parameter utilisation, and task performance across diverse NLP tasks, while also fostering interpretable expert specialisation. Its efficiency, aided by fused kernels, makes it a promising direction for scaling transformers. Future work may explore hierarchical routing combining slice- and token-level decisions, adaptive slice counts, and porting fused kernels to a wider range of emerging accelerators.

## Limitations

SliceMoE, while promising, has several limitations and areas for future investigation:

**Scalability** The router MLP's input dimension is $d/S$. While lightweight for moderate $d$ and $S$, the total routing FLOPs scale with $S$ (number of slices per token). For extremely large $S$ or a very high number of experts $E$, routing computation could become a bottleneck relative to expert computation. Hierarchical routing or dynamically determined slice counts could mitigate this.

**Hyperparameter Sensitivity** The number of slices $S$, top-$k$ expert choices, capacity loss weight $\alpha$, and cross-slice dropout rate are crucial hyperparameters requiring careful tuning for optimal performance. The ideal $S$ may also depend on the embedding dimension $d$ and the specific task.

**Hardware Dependency for Fused Kernels** The reported efficiency gains rely on fused batched

GEMM kernels, which are most effective on modern GPUs like A100s. Performance benefits might be less pronounced on older hardware or if less optimized kernel implementations are used. Broader hardware compatibility and optimized open-source kernels would enhance practical adoption.

**Classification Experimental Setup** Our classification experiments utilized a frozen DistilBERT encoder to isolate the MoE layer's impact. While this allows for a focused comparison of routing strategies, these results may not directly generalize to scenarios involving full end-to-end fine-tuning of the entire model. Exploring SliceMoE in fully trainable large models is an important next step.

**Comparisons with SOTA MoE Variants** While SliceMoE demonstrates significant improvements over standard token-level MoE and dense baselines, this work did not include exhaustive comparisons against all recent, highly specialized MoE architectures (e.g., those with very complex learned routing or dynamic expert merging/pruning). Such comparisons would provide a more complete picture of SliceMoE's relative standing.

**Increased Implementation Complexity** Slice-level routing and aggregation introduce more routing decisions and data manipulation steps compared to token-level routing, potentially increasing the initial implementation complexity.

**Interpretability Metrics** Our current interpretability analysis, while indicative, relies on specific metrics like ESS and qualitative examples. Developing more comprehensive and standardized quantitative metrics for expert specialization in MoE models remains an open research area.

## References

Zexuan Chen, Zihui Li, Xuan Zhao, A-Long Zhou, Lichao Yu, Wei Zhang, and Ming Zhou. 2023. PR-MoE: Post-routing mixture of experts. In *Proceedings of the 61st Annual Meeting of the Association for Computational Linguistics (Volume 1: Long Papers)*.

Zixiang Chen, Yihe Deng, Yue Wu, Quanquan Gu, and Yuan-Fang Li. 2022. Towards understanding the mixture-of-experts layer in deep learning. *Advances in Neural Information Processing Systems*, 35:3136–3149.

I. M. Chowdhury, Kai Su, and Qiangfu Zhao. 2020. MS-NET: Modular selective network. *International Journal of Machine Learning and Cybernetics*, 12:763–781.

W. Fedus, B. Zoph, and N. Shazeer. 2021. Switch transformers: Scaling to trillion parameter models with simple and efficient sparsity. *arXiv preprint arXiv:2101.03961*.

Ethan He, Abhinav Khattar, Ryan Prenger, Oleksii Kuchaiev, Anima Liu, and Boris Ginsburg.

Rui Kong, Qiyang Li, Xinyu Fang, Haotian Chen, Guohao Zhao, Guangtou Zhao, Yuchen Wang, Zhen Cheng, Ming Zhang, Wen Xiao, and Yu Wang.

Mohammed Muqeeth, Haokun Liu, and Colin Raffel. 2023. Soft merging of experts with adaptive routing. *arXiv preprint arXiv:2306.03745*.

Imrus Salehin and Dae-Ki Kang. 2023. A review on dropout regularization approaches for deep neural networks. *Electronics*, 12(5).

Liang Shen, Zhihua Wu, Weibao Gong, Hongxiang Hao, Yangfan Bai, Huachao Wu, Xinxuan Wu, Haoyi Xiong, Dianhai Yu, and Yanjun Ma. 2022. MoESys: A distributed and efficient mixture-of-experts training and inference system for internet services. *IEEE Transactions on Services Computing*, 17:2626–2639.

Sandeep Subramanian, Oleksii Hrinchuk, Virginia Adams, and Oleksii Kuchaiev. NVIDIA NeMo's neural machine translation systems at WMT21. In *Proceedings of the Sixth Conference on Machine Translation (WMT21)*.

Luyu Wang, Yujia Li, Özlem Aslan, and Oriol Vinyals. WikiGraphs: A wikipedia text–knowledge graph paired dataset. In *Proceedings of the 15th Workshop on Graph-Based Methods for Natural Language Processing (TextGraphs-15)*.

Shunhong Wang. FlexGEMM: A flexible micro-kernel generation framework.

## A Hyper-parameter Details

The following details supplement Section 4:

- Transformer hidden dimension $d = 768$, FFN intermediate dimension $4 \times d = 3072$.
- SliceMoE Router MLP: Input $d/S$, hidden layer $H_r = 256$ with ReLU, output $E$ (number of experts). For $S = 8$, $d/S = 96$.
- Number of Slices $S$: Varied in $\{2, 4, 8, 16, 32\}$ for ablation (Table 2). $S = 8$ was generally optimal.
- Top-$k$ experts per slice: $k = 2$ used consistently.
- Capacity loss weight $\alpha$: Validated in range [0.01, 0.2], set to 0.1 for LM/MT and 0.05 for classification for best stability and ELE.

- Cross-slice dropout rate: 0.2 (i.e., 20% of selected expert assignments per slice dropped). Standard dropout of 0.1 on FFN activations.

## B Additional Ablation Results

Figure 7 shows the effect of varying the softmax temperature in the slice router on AG NEWS validation accuracy. Performance is relatively stable for temperatures between 0.5 and 2.0, with a slight peak around 1.0 (default used).

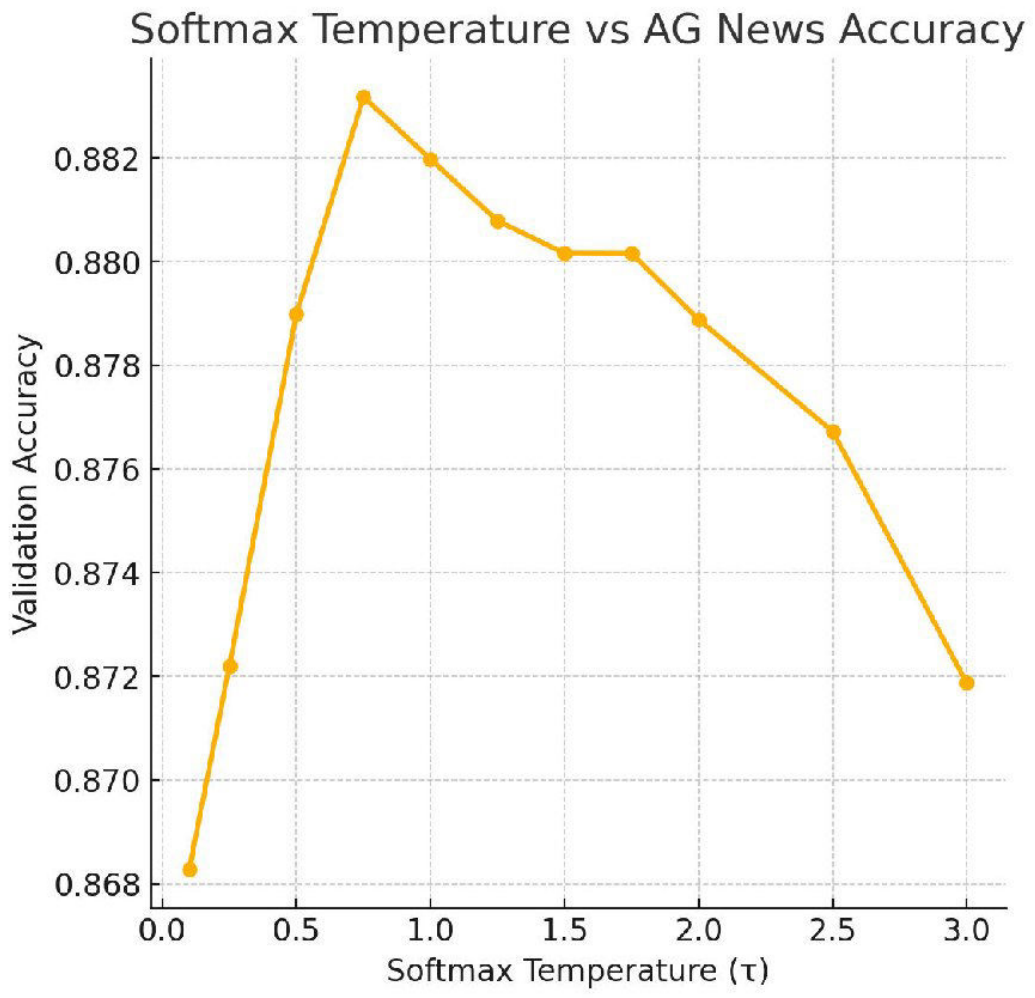

Figure 7: Effect of router softmax temperature on AG NEWS validation accuracy for SliceMoE (S=8, k=2).

**Impact of Experts per Slice (k)** We ablate the number of experts selected per slice ($k$) on AG NEWS, shown in Table 4. Using $k = 1$ is fastest but results in lower accuracy and balance, as it limits the router's flexibility. Increasing from $k = 2$ to $k = 3$ offers a marginal accuracy improvement at a significant latency cost. This confirms that $k = 2$ provides a robust trade-off between performance and efficiency for our setup.

| $k$ | Accuracy ↑ | ELE ↑ | Latency (ms/batch) ↓ |
|---|---|---|---|
| 1 | 0.874 | 0.91 | **15.2** |
| **2** | **0.880** | **0.95** | 18.5 |
| 3 | 0.881 | 0.94 | 24.1 |

Table 4: Ablation on the number of experts per slice ($k$) for SliceMoE (S=8) on AG NEWS. $k = 2$ offers the best balance.

## C Additional Figures and Interpretability Examples

Figure 8 shows the confusion matrix for SliceMoE on DBPEDIA-14, indicating strong performance across most classes.

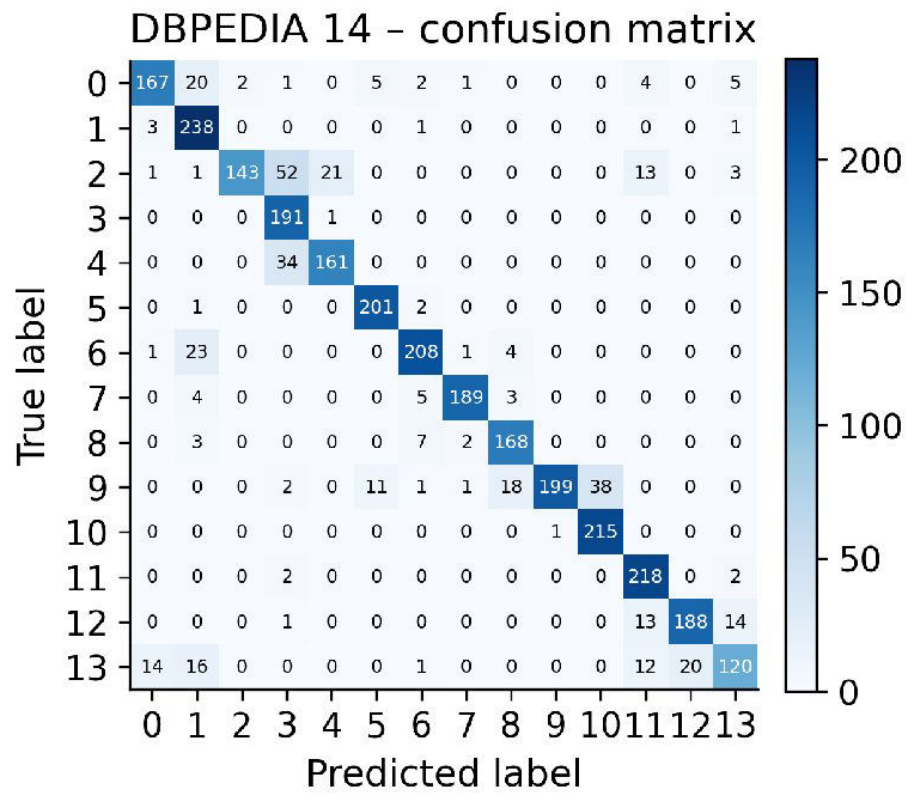

Figure 8: Confusion matrix for DBPEDIA-14 (SliceMoE, S=8), showing improved class-wise performance compared to TokenMoE (not shown).

Figure 9 illustrates SliceMoE's behavior on a synthetic task designed with distinct features in different embedding segments. SliceMoE quickly learns to route corresponding slices to specialized experts, achieving near-perfect load balance.

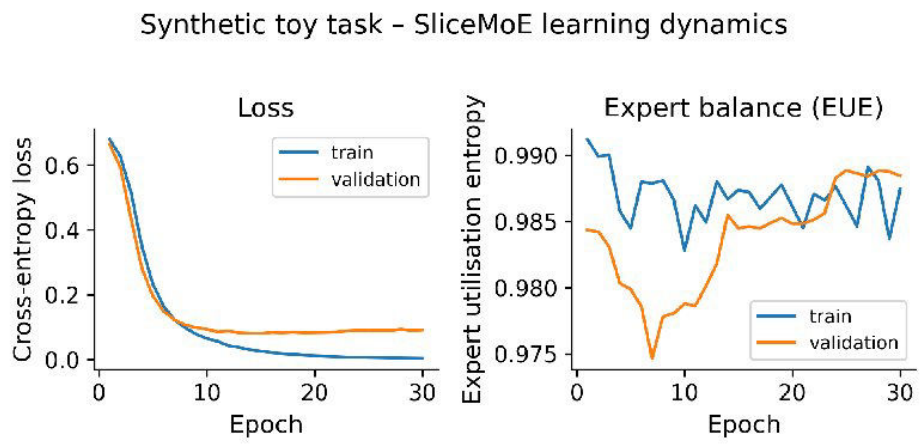

Figure 9: SliceMoE on a synthetic toy task: expert load entropy (ELE) rapidly converges to near-optimal balance within five epochs.

**Further Interpretability Examples (AG NEWS, S=8):**

- **Expert 3 (Financial/Business):** High activation for slices from tokens/phrases like "Inc.", "Corp.", "stocks fell", "quarterly results", "market share". Input slice embeddings show tighter clustering around business concepts.
- **Expert 7 (Technology/Science):** High activation for slices from "software", "version", "internet", "researchers", "nasa".
- **Expert 12 (World Affairs/Politics):** High activation for slices from "government", "election", "minister", "United Nations", "conflict".

These qualitative observations, alongside the ESS metric, reinforce the finding that SliceMoE experts develop more granular specializations.

**Generalizability of Slicing Assumption (WMT En-De):** To test if our slicing assumption holds on other tasks, we performed a preliminary analysis on WMT En-De translation. We found evidence of specialization here as well. One expert consistently received slices from early-to-mid vector indices (e.g., dimensions 128-256), which we found correlated with source-side verb tense information. Another expert showed high activation for slices from later indices (e.g., dimensions 512-640), which correlated with noun phrases and gender agreement cues. This suggests the principle of local, coherent information within embedding vectors is not task-specific.